\documentclass[11pt,conference]{IEEEtran}
\IEEEoverridecommandlockouts
\usepackage{cite}
\usepackage{amsmath,amssymb,amsfonts}
\usepackage{graphicx}
\usepackage{textcomp}
\usepackage{xcolor}
\usepackage{booktabs}
\usepackage{multirow}

\def\BibTeX{{\rm B\kern-.05em{\sc i\kern-.025em b}\kern-.08em
    T\kern-.1667em\lower.7ex\hbox{E}\kern-.125emX}}
\begin{document}

\title{MSCA-UNet: Multi-Scale Context and Attention U-Net for Image Segmentation}

\author{
\IEEEauthorblockN{Sheng-Wei Chan}
\IEEEauthorblockA{
\textit{Dept. of Electrical Engineering} \\
\textit{Tamkang University} \\
412440330@o365.tku.edu.tw
}
}

\maketitle

\begin{abstract}
U-Net remains a practical baseline for image segmentation because of its simple encoder--decoder structure and skip connections. However, the bottleneck representation is still dominated by a limited set of receptive fields, while decoder features are propagated without explicitly emphasizing the most informative channels and spatial locations. This paper presents MSCA-UNet, a U-Net-based segmentation architecture that combines multi-scale contextual aggregation at the bottleneck with channel-spatial attention refinement in the decoder. The multi-scale module uses parallel atrous convolutions to capture contextual features at different receptive fields, while Convolutional Block Attention Modules (CBAMs) progressively recalibrate decoder features. Under identical experimental settings, the baseline U-Net achieves 96.9\% mIoU on a held-out test set. Adding multi-scale context improves mIoU to 97.5\%, while attention alone reaches 98.4\%. Combining both mechanisms yields 99.1\% mIoU, a 2.2 percentage-point improvement over the baseline. Parameter analysis further shows that the attention-only variant adds approximately 0.044M parameters, whereas the multi-scale module contributes most of the additional model capacity. The results support the view that multi-scale context enrichment and attention-based feature refinement provide complementary benefits within a U-Net framework.
\end{abstract}


\section{Introduction}
Image segmentation is a fundamental computer vision task that assigns a semantic label to each pixel and supports applications in autonomous systems, industrial inspection, medical imaging, and scene understanding. Fully convolutional networks established the modern end-to-end formulation of dense prediction~\cite{long2015fully}, while U-Net introduced a symmetric encoder--decoder architecture with skip connections that effectively combines high-level semantic information and low-level spatial detail~\cite{ronneberger2015unet}. Owing to its implementation simplicity and strong localization capability, U-Net continues to serve as a useful baseline for binary and multi-class segmentation.

Despite its effectiveness, two limitations remain relevant. First, the standard bottleneck mainly aggregates information through fixed local convolutional kernels. Objects can appear at different spatial scales, so explicitly modeling multiple receptive fields can enrich the high-level representation. Atrous convolution and atrous spatial pyramid pooling have demonstrated that multi-scale context can improve dense prediction while preserving spatial resolution~\cite{chen2017deeplab,chen2017rethinking,chen2018encoder}. Second, decoder features contain both useful and redundant responses. Attention mechanisms provide a lightweight way to recalibrate these features by emphasizing informative channels and spatial locations. CBAM, in particular, sequentially applies channel and spatial attention with relatively small parameter overhead~\cite{woo2018cbam}.

These observations motivate a practical question: \emph{are multi-scale contextual modeling and attention-based feature refinement redundant, or do they provide complementary improvements when integrated into U-Net?} To study this question, we construct MSCA-UNet, which places a parallel atrous context module after the U-Net bottleneck and applies CBAM after each decoder convolution block. We evaluate four controlled configurations under the same experimental protocol: the baseline U-Net, U-Net with multi-scale context only, U-Net with attention only, and the complete MSCA-UNet.

The main contributions of this work are summarized as follows:
\begin{itemize}
    \item We present MSCA-UNet, a straightforward U-Net extension that combines multi-scale contextual aggregation at the bottleneck with channel-spatial feature refinement throughout the decoder.
    \item We conduct a controlled component-wise ablation study to quantify the individual and joint effects of multi-scale context and attention. Both components independently improve segmentation accuracy, and their combination provides the highest mIoU.
    \item We analyze the accuracy--complexity trade-off of the two mechanisms. The attention-only variant adds only 43,912 trainable parameters (about 0.14\% over the baseline) while improving mIoU by 1.5 percentage points, whereas the complete model reaches 99.1\% mIoU.
\end{itemize}

\section{Related Work}
\subsection{Encoder--Decoder Segmentation}
FCN formulated semantic segmentation as end-to-end dense prediction using fully convolutional networks~\cite{long2015fully}. U-Net further introduced a symmetric encoder--decoder design with skip connections that recover fine spatial detail from early encoder stages~\cite{ronneberger2015unet}. The architecture has since become a widely used baseline because it offers a favorable balance between implementation simplicity and segmentation quality. Our work retains the U-Net hierarchy and studies how its bottleneck and decoder representations can be strengthened without replacing the overall encoder--decoder structure.

\subsection{Multi-Scale Context Modeling}
Multi-scale context is important for segmentation because objects and structures can occupy substantially different spatial extents. DeepLab introduced atrous convolution for enlarging the effective receptive field while maintaining feature resolution, and later variants employed atrous spatial pyramid pooling to aggregate contextual information at multiple dilation rates~\cite{chen2017deeplab,chen2017rethinking,chen2018encoder}. Pyramid pooling provides a related strategy for collecting context at several spatial scales~\cite{zhao2017pyramid}. MSCA-UNet follows this general principle by applying parallel atrous branches to the bottleneck feature map before decoding.

\subsection{Attention-Based Feature Refinement}
Channel attention methods such as Squeeze-and-Excitation Networks reweight feature channels according to global statistics~\cite{hu2018senet}. Attention U-Net demonstrated the usefulness of attention gating within an encoder--decoder segmentation framework~\cite{oktay2018attention}. CBAM extends channel recalibration by sequentially applying channel and spatial attention, enabling a network to emphasize both \emph{what} and \emph{where} informative features are located~\cite{woo2018cbam}. In MSCA-UNet, CBAM is inserted after each decoder convolution block so that reconstructed features are progressively refined at multiple resolutions.

\section{MSCA-UNet}
\subsection{Overall Architecture}
MSCA-UNet follows a four-stage U-Net topology. The encoder uses feature widths of 64, 128, 256, and 512 channels, followed by a 1024-channel bottleneck. Each encoder stage contains two $3\times3$ convolutions with batch normalization and ReLU activation, and $2\times2$ max pooling performs downsampling. The decoder uses $2\times2$ transposed convolutions for upsampling, concatenates the corresponding encoder skip feature, and applies another two-convolution block. A final $1\times1$ convolution produces the binary segmentation logit map.

\begin{figure}[t]
\centering
\includegraphics[width=0.84\linewidth]{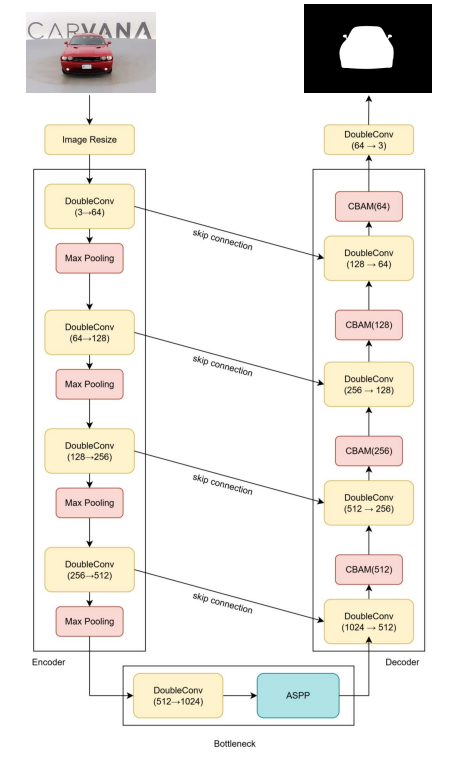}
\caption{Architecture of MSCA-UNet. An ASPP-style multi-scale context module is applied at the bottleneck, while CBAM is inserted after each decoder convolution block for channel-spatial feature refinement.}
\label{fig:architecture}
\end{figure}

The complete architecture introduces two modifications. First, the bottleneck output is processed by an ASPP-style multi-scale context module that aggregates features from several dilation rates. Second, a CBAM block is applied after each decoder convolution block. Fig.~\ref{fig:architecture} shows the implemented network.

\subsection{Multi-Scale Context Module}
Let $F_b\in\mathbb{R}^{H_b\times W_b\times C_b}$ denote the bottleneck feature map, where $C_b=1024$ in the implemented model. To enrich the representation with different receptive fields, four parallel branches are applied. One branch uses a $1\times1$ convolution, while three branches use $3\times3$ atrous convolutions with dilation rates $d\in\{3,6,12\}$:
\begin{equation}
F_0 = \mathcal{C}_{1\times1}(F_b),
\end{equation}
\begin{equation}
F_i = \mathcal{C}^{d_i}_{3\times3}(F_b), \quad d_i\in\{3,6,12\}.
\end{equation}
The branch outputs are concatenated along the channel dimension and projected back to the bottleneck width:
\begin{equation}
F_{ms}=\rho\!\left(\mathrm{BN}\!\left(\mathcal{C}_{1\times1}\!\left(\mathrm{BN}\!\left([F_0,F_1,F_2,F_3]\right)\right)\right)\right),
\end{equation}
where $[\cdot]$ denotes channel-wise concatenation, BN denotes batch normalization, and $\rho(\cdot)$ is ReLU. The resulting feature $F_{ms}$ replaces the original bottleneck output as the decoder input. Because every branch preserves the full bottleneck width before concatenation, this module is intentionally high-capacity and accounts for most of the additional parameters in MSCA-UNet.

\subsection{Channel-Spatial Attention Refinement}
CBAM is applied after every decoder convolution block. For a decoder feature $F\in\mathbb{R}^{H\times W\times C}$, channel attention first aggregates spatial information using global average pooling and global max pooling. The two descriptors are processed by a shared two-layer multilayer perceptron with reduction ratio $r=16$:
\begin{equation}
M_c(F)=\sigma\left(\mathrm{MLP}(\mathrm{GAP}(F))+\mathrm{MLP}(\mathrm{GMP}(F))\right),
\end{equation}
where $\sigma(\cdot)$ is the sigmoid function. Channel-refined features are then obtained by
\begin{equation}
F_c=M_c(F)\odot F.
\end{equation}
Spatial attention summarizes $F_c$ across channels using average and maximum operations and applies a $7\times7$ convolution:
\begin{equation}
M_s(F_c)=\sigma\left(\mathcal{C}_{7\times7}([\mathrm{Avg}_c(F_c),\mathrm{Max}_c(F_c)])\right).
\end{equation}
The final refined decoder feature is
\begin{equation}
F_a=M_s(F_c)\odot F_c.
\end{equation}
This sequence enables the decoder to adaptively emphasize informative semantic channels and spatial locations at each reconstruction scale.

\subsection{Training Objective}
All variants are optimized using the same weighted combination of binary cross-entropy, Dice loss, and a boundary-aware loss:
\begin{equation}
\mathcal{L}=0.45\mathcal{L}_{BCE}+0.30\mathcal{L}_{Dice}+0.25\mathcal{L}_{Bnd}.
\end{equation}
The Dice term encourages region-level overlap between the prediction and target mask. The boundary term applies horizontal and vertical Sobel filters to the sigmoid prediction and ground-truth mask and minimizes the $L_1$ distance between their gradient responses. The objective therefore combines pixel-wise classification, region overlap, and boundary consistency.

\section{Experiments}
\subsection{Dataset and Preprocessing}
Experiments are conducted on vehicle foreground segmentation data derived from the Carvana Image Masking Challenge~\cite{carvana2017}. The project subset contains 1,600 RGB vehicle images with corresponding binary masks. Images and masks are resized to $384\times384$ pixels, and nearest-neighbor interpolation is used for masks to preserve binary labels. A fixed held-out test partition is reserved for final evaluation and is not used for model optimization. The same data partition is used for all four ablation configurations.

\subsection{Implementation Details}
All four configurations use the same U-Net channel widths $(64,128,256,512)$ and the same training protocol. The input resolution is $384\times384$ with a physical batch size of 2. Training is performed with Adam in three consecutive 10-epoch stages. The initial learning rates are $1\times10^{-4}$, $5\times10^{-5}$, and $2\times10^{-5}$ for the three stages, respectively. Within each stage, StepLR reduces the learning rate by a factor of 0.5 every five epochs. The mixed objective described above is used throughout training. Experiments are conducted in a Windows environment with an NVIDIA GeForce RTX 3080 Ti GPU.

\subsection{Evaluation Metric}
Intersection over Union (IoU) for class $c$ is defined as
\begin{equation}
\mathrm{IoU}_c=\frac{TP_c}{TP_c+FP_c+FN_c},
\end{equation}
and mean IoU is
\begin{equation}
\mathrm{mIoU}=\frac{1}{C}\sum_{c=1}^{C}\mathrm{IoU}_c,
\end{equation}
where $C$ is the number of evaluated classes. We additionally report the number of trainable parameters to examine the accuracy--complexity trade-off.

\section{Results and Discussion}
\subsection{Component-Wise Ablation}
Table~\ref{tab:ablation} compares the four controlled configurations. The baseline U-Net achieves 96.9\% mIoU. Adding only the multi-scale context module improves mIoU to 97.5\%, corresponding to a gain of 0.6 percentage points. Adding only CBAM improves mIoU to 98.4\%, a gain of 1.5 points. When both components are enabled, MSCA-UNet reaches 99.1\% mIoU, outperforming the baseline by 2.2 percentage points.

\begin{table}[t]
\caption{Component-wise ablation of MSCA-UNet on the held-out test set.}
\label{tab:ablation}
\centering
\scriptsize
\setlength{\tabcolsep}{3.0pt}
\begin{tabular}{lcccc}
\toprule
\textbf{Model} & \textbf{Multi-Scale} & \textbf{CBAM} & \textbf{Params. (M)} & \textbf{mIoU (\%)} \\
\midrule
U-Net & -- & -- & 31.04 & 96.9 \\
U-Net + Multi-Scale & \checkmark & -- & 64.60 & 97.5 \\
U-Net + Attention & -- & \checkmark & 31.08 & 98.4 \\
\textbf{MSCA-UNet} & \checkmark & \checkmark & \textbf{64.65} & \textbf{99.1} \\
\bottomrule
\end{tabular}
\end{table}

The ablation results support two observations. First, multi-scale contextual aggregation provides a measurable improvement by exposing the bottleneck representation to multiple receptive fields. Second, attention refinement yields a larger individual gain while introducing very limited parameter growth. Most importantly, the complete model outperforms either individual component, suggesting that the two mechanisms remain complementary rather than redundant: multi-scale context enriches the available representation, whereas attention selectively refines the information propagated through the decoder.

\subsection{Accuracy--Complexity Trade-off}
The baseline U-Net contains 31,037,633 trainable parameters. The attention-only variant contains 31,081,545 parameters, adding 43,912 parameters (approximately 0.044M, or 0.14\%) while improving mIoU by 1.5 percentage points. By contrast, the multi-scale-only variant contains 64,602,305 parameters because each parallel bottleneck branch preserves the full 1024-channel width before concatenation and projection. The complete MSCA-UNet contains 64,646,217 parameters.

These results highlight different roles for the two mechanisms. CBAM provides the stronger individual accuracy gain while adding only 0.14\% more parameters, making it substantially more parameter-efficient than the multi-scale context module. The multi-scale module introduces considerably more representational capacity and provides a smaller standalone gain, but it remains complementary to attention: adding multi-scale context to the attention-enhanced model further improves mIoU from 98.4\% to 99.1\%. The complete model therefore achieves the highest test accuracy, whereas the attention-only configuration offers a favorable alternative when model size is more constrained.

\subsection{Representative Prediction Results}
\begin{figure*}[t]
\centering

\begin{minipage}[t]{0.315\textwidth}
\centering
\includegraphics[width=\linewidth,height=1.18in,keepaspectratio]{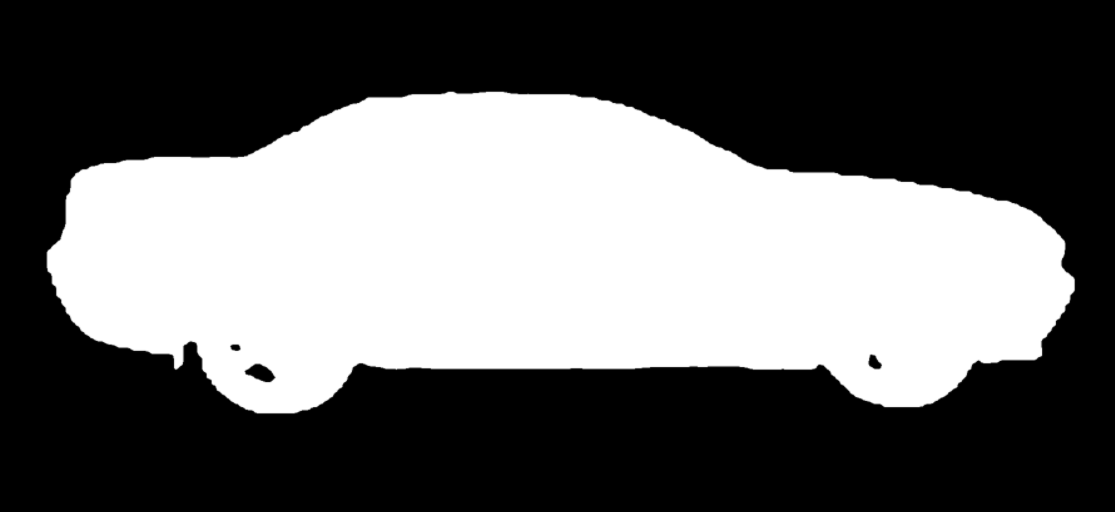}\\[-1pt]
\footnotesize (a)
\end{minipage}\hfill
\begin{minipage}[t]{0.315\textwidth}
\centering
\includegraphics[width=\linewidth,height=1.18in,keepaspectratio]{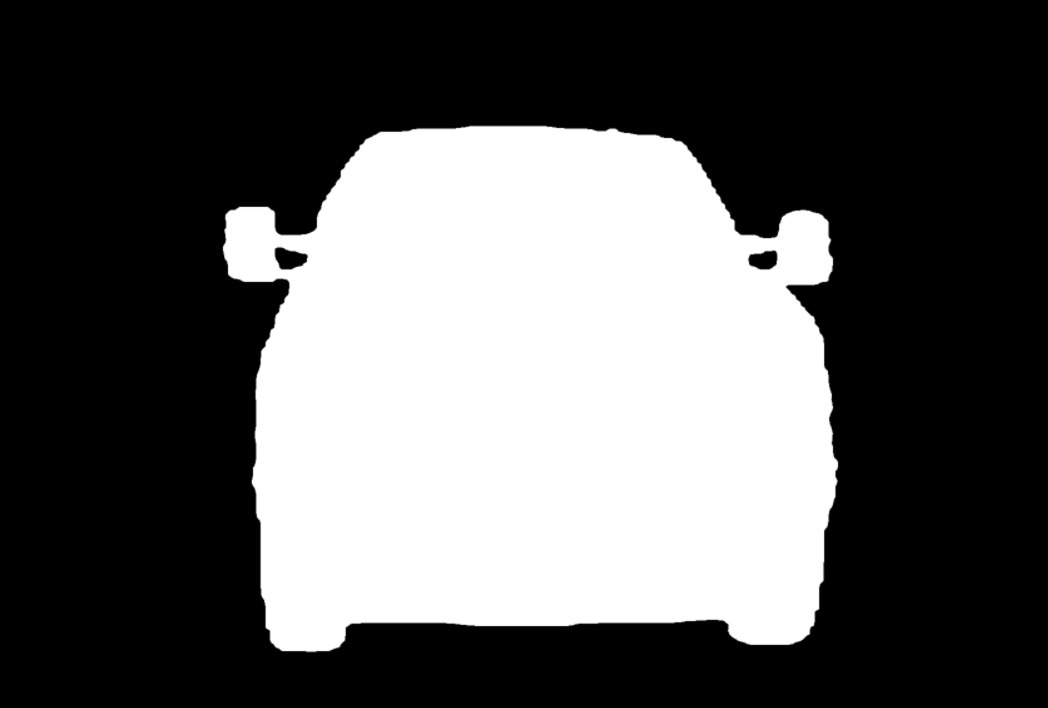}\\[-1pt]
\footnotesize (b)
\end{minipage}\hfill
\begin{minipage}[t]{0.315\textwidth}
\centering
\includegraphics[width=\linewidth,height=1.18in,keepaspectratio]{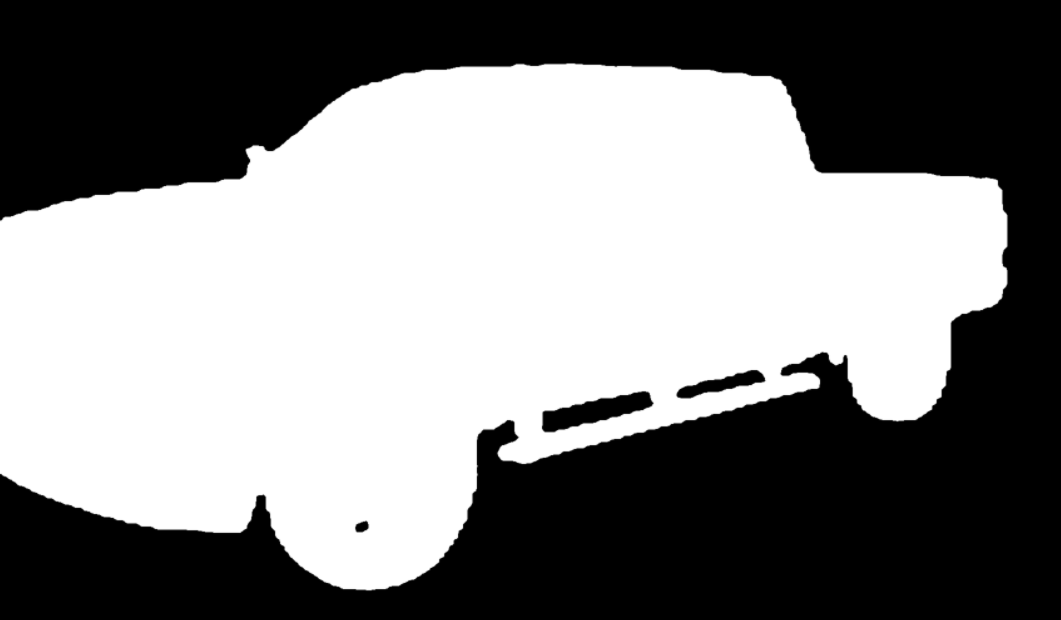}\\[-1pt]
\footnotesize (c)
\end{minipage}

\vspace{5pt}

\begin{minipage}[t]{0.315\textwidth}
\centering
\includegraphics[width=\linewidth,height=1.18in,keepaspectratio]{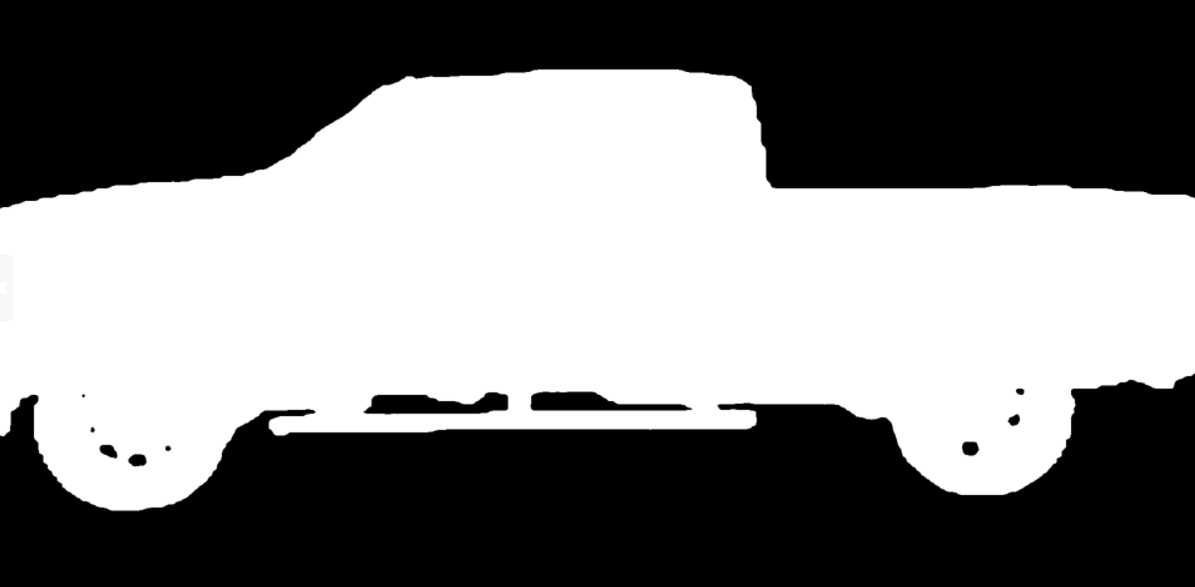}\\[-1pt]
\footnotesize (d)
\end{minipage}\hfill
\begin{minipage}[t]{0.315\textwidth}
\centering
\includegraphics[width=\linewidth,height=1.18in,keepaspectratio]{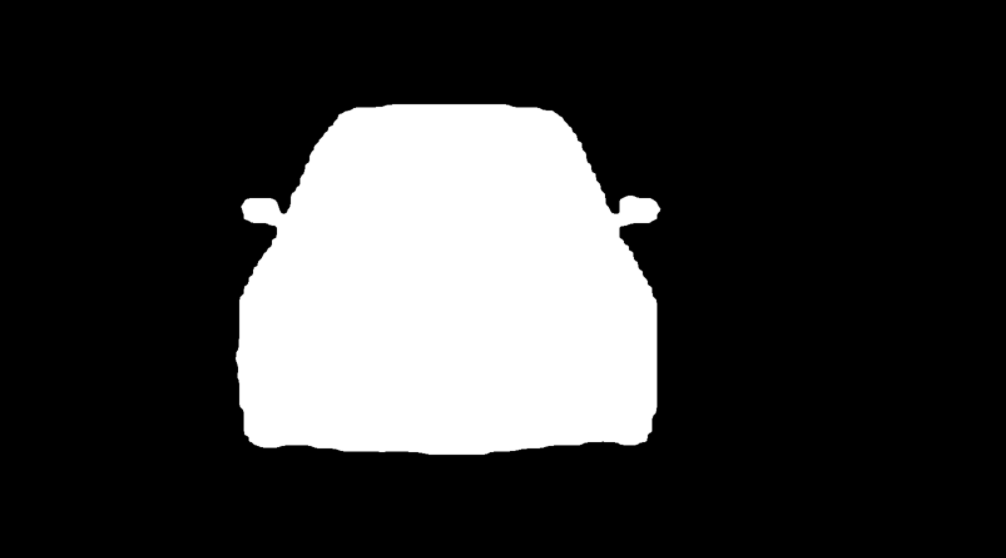}\\[-1pt]
\footnotesize (e)
\end{minipage}\hfill
\begin{minipage}[t]{0.315\textwidth}
\centering
\includegraphics[width=\linewidth,height=1.18in,keepaspectratio]{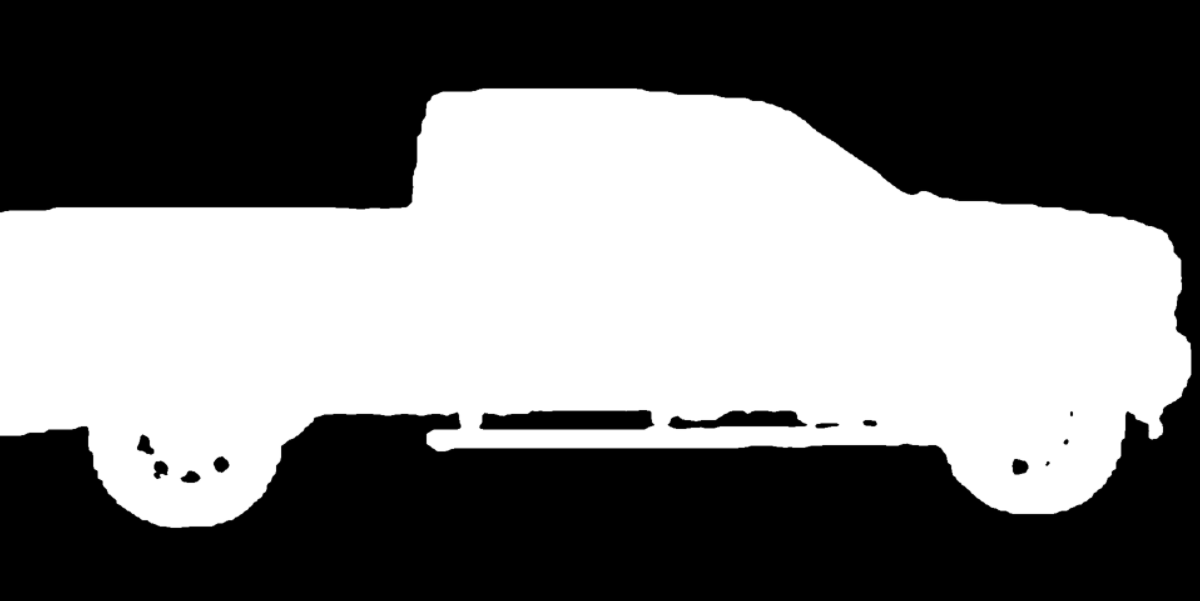}\\[-1pt]
\footnotesize (f)
\end{minipage}

\caption{Representative binary segmentation predictions produced by MSCA-UNet. Six examples are arranged in a $2\times3$ layout to illustrate different vehicle orientations and body shapes. The masks preserve coherent global foreground regions across diverse viewpoints, while fine structures around the underbody and wheel regions show greater variation.}
\label{fig:qualitative}
\end{figure*}

Fig.~\ref{fig:qualitative} presents six representative binary masks generated by MSCA-UNet for vehicles with different viewpoints and body shapes. The predictions preserve coherent global vehicle silhouettes across frontal, lateral, and oblique views, while also retaining smaller structures such as side mirrors and wheel regions in several examples. Fine structures around the underbody and wheel regions are less consistently delineated than the overall vehicle silhouette, suggesting that fine-detail recovery remains more challenging than global foreground localization.

\subsection{Limitations}
The current study evaluates the architecture on a single vehicle segmentation dataset and uses a relatively high-capacity multi-scale bottleneck. Consequently, the experiments establish the component behavior under the present setting rather than universal superiority across segmentation tasks. In particular, the multi-scale module approximately doubles the parameter count relative to the baseline, and this study does not report FLOPs or measured inference latency. Future work should investigate narrower multi-scale branches, additional datasets, and deployment-oriented efficiency measurements to determine whether the observed complementarity generalizes to different object categories and resource constraints.

\section{Conclusion}
This paper presented MSCA-UNet, a U-Net-based segmentation architecture that combines multi-scale contextual aggregation with channel-spatial attention refinement. Controlled ablation experiments show that the multi-scale and attention components independently improve the baseline from 96.9\% mIoU to 97.5\% and 98.4\%, respectively. Combining both mechanisms achieves 99.1\% mIoU, a 2.2 percentage-point improvement over the baseline. Parameter analysis further shows that CBAM provides the larger individual improvement with negligible overhead, whereas the multi-scale bottleneck is responsible for most of the additional model capacity. The further improvement from 98.4\% to 99.1\% when multi-scale context is added to the attention-enhanced model supports the view that the two mechanisms address different aspects of representation learning and can be effectively combined within U-Net. Future work will investigate lighter multi-scale designs and validation on additional segmentation datasets.


\begin{thebibliography}{00}
\bibitem{long2015fully} J. Long, E. Shelhamer, and T. Darrell, ``Fully convolutional networks for semantic segmentation,'' in \emph{Proc. IEEE Conf. Comput. Vis. Pattern Recognit. (CVPR)}, 2015, pp. 3431--3440.
\bibitem{ronneberger2015unet} O. Ronneberger, P. Fischer, and T. Brox, ``U-Net: Convolutional networks for biomedical image segmentation,'' in \emph{Proc. Med. Image Comput. Comput.-Assist. Intervent. (MICCAI)}, 2015, pp. 234--241.
\bibitem{chen2017deeplab} L.-C. Chen, G. Papandreou, I. Kokkinos, K. Murphy, and A. L. Yuille, ``DeepLab: Semantic image segmentation with deep convolutional nets, atrous convolution, and fully connected CRFs,'' \emph{IEEE Trans. Pattern Anal. Mach. Intell.}, vol. 40, no. 4, pp. 834--848, 2018.
\bibitem{chen2017rethinking} L.-C. Chen, G. Papandreou, F. Schroff, and H. Adam, ``Rethinking atrous convolution for semantic image segmentation,'' arXiv:1706.05587, 2017.
\bibitem{chen2018encoder} L.-C. Chen, Y. Zhu, G. Papandreou, F. Schroff, and H. Adam, ``Encoder-decoder with atrous separable convolution for semantic image segmentation,'' in \emph{Proc. Eur. Conf. Comput. Vis. (ECCV)}, 2018, pp. 801--818.
\bibitem{zhao2017pyramid} H. Zhao, J. Shi, X. Qi, X. Wang, and J. Jia, ``Pyramid scene parsing network,'' in \emph{Proc. IEEE Conf. Comput. Vis. Pattern Recognit. (CVPR)}, 2017, pp. 2881--2890.
\bibitem{hu2018senet} J. Hu, L. Shen, and G. Sun, ``Squeeze-and-excitation networks,'' in \emph{Proc. IEEE Conf. Comput. Vis. Pattern Recognit. (CVPR)}, 2018, pp. 7132--7141.
\bibitem{oktay2018attention} O. Oktay \emph{et al.}, ``Attention U-Net: Learning where to look for the pancreas,'' arXiv:1804.03999, 2018.
\bibitem{woo2018cbam} S. Woo, J. Park, J.-Y. Lee, and I. S. Kweon, ``CBAM: Convolutional block attention module,'' in \emph{Proc. Eur. Conf. Comput. Vis. (ECCV)}, 2018, pp. 3--19.
\bibitem{carvana2017} Carvana, ``Carvana Image Masking Challenge,'' Kaggle competition dataset, 2017.
\end{thebibliography}
\end{document}